\documentclass[runningheads]{llncs}
\usepackage{graphicx}
\usepackage{amsmath}
\usepackage{color}
\usepackage{soul}
\usepackage{multirow}
\usepackage{amssymb}
\usepackage{wrapfig}
\usepackage{mathtools}
\usepackage[ruled,linesnumbered]{algorithm2e}
\usepackage{esvect}
\usepackage{ textcomp }
\usepackage{threeparttable}
\usepackage{amssymb}
\usepackage{float}
\usepackage{subfigure}
\usepackage[pagebackref,breaklinks,colorlinks]{hyperref}
\usepackage{cite}

\SetKwRepeat{Do}{do}{while}%

\definecolor{turquoise}{cmyk}{0.65,0,0.1,0.1}
\definecolor{purple}{rgb}{0.65,0,0.65}
\definecolor{darkgreen}{rgb}{0.0, 0.5, 0.0}
\definecolor{darkred}{rgb}{0.5, 0.0, 0.0}
\definecolor{darkblue}{rgb}{0.0, 0.0, 0.5}
\definecolor{myblue}{rgb}{0.0, 0.0, 1.0}
\definecolor{magenta}{rgb}{1.0,0,1.0}
\definecolor{myred}{rgb}{0.9, 0.0, 0.0}

\newcommand{\erase}[1]{}

\newcommand{\hide}[1]{{\color{red}-- -- -- --}}

\definecolor{blue}{rgb}{0.0, 0.0, 1.0}

\usepackage{tikz}
\usepackage{comment}
\usepackage{amsmath,amssymb} 
\usepackage{color}
\usepackage{cite}

\graphicspath{{figs/}}

\begin{document}
\pagestyle{headings}
\mainmatter
\def\ECCVSubNumber{5641}  

\title{Deep Point Cloud Simplification for High-quality Surface Reconstruction} 


\titlerunning{Deep point simplification}
%
\author{Yuanqi Li\inst{1}\orcidID{0000-0003-4100-7471} \and
Jianwei Guo\inst{2} \and
Xinran Yang\inst{1} \and
Shun Liu\inst{1} \and
Jie Guo\inst{1} \and
Xiaopeng Zhang \inst{2} \and
Yanwen Guo \inst{1}
}
\authorrunning{Y. Li et al.}
%
\institute{National Key Lab for Novel Software Technology, Nanjing University\and
NLPR, Institute of Automation, Chinese Academy of Sciences}
\maketitle

\begin{abstract}
The growing size of point clouds enlarges consumptions of storage, transmission, and computation of 3D scenes. Raw data is redundant, noisy, and non-uniform. Therefore, simplifying point clouds for achieving compact, clean, and uniform points is becoming increasingly important for 3D vision and graphics tasks.
Previous learning based methods aim to generate fewer points for scene understanding, regardless of the quality of surface reconstruction, leading to results with low reconstruction accuracy and bad point distribution.  
In this paper, we propose a novel point cloud simplification network (PCS-Net) dedicated to high-quality surface mesh reconstruction while maintaining geometric fidelity. 
We first learn a sampling matrix in a \emph{feature-aware simplification module} to reduce the number of points.
Then we propose a novel \emph{double-scale resampling module} to refine the positions of the sampled points, to achieve a uniform distribution. 
To further retain important shape features, an adaptive sampling strategy with a novel saliency loss is designed.
With our PCS-Net, the input non-uniform and noisy point cloud can be simplified in a feature-aware manner, i.e., points near salient features are consolidated but still with uniform distribution locally. Experiments demonstrate the effectiveness of our method and show that we outperform previous simplification or reconstruction-oriented upsampling methods. 

\end{abstract}

\section{Introduction}\label{sec:introduction}
\begin{figure}[tb]
\centering
\includegraphics[width=.98\linewidth]{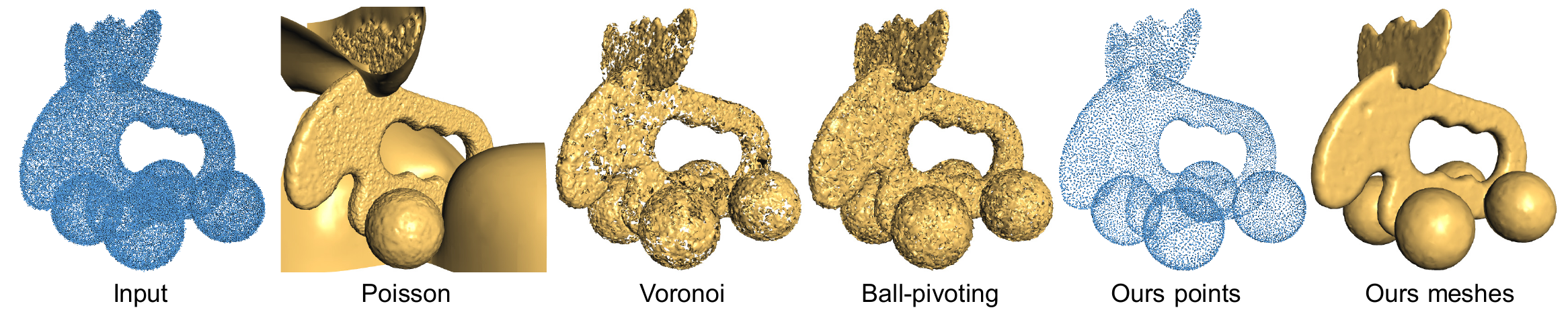}
\parbox[t]{.98\columnwidth}{\relax
}
\caption{\label{reconstruct}
An illustration of direct surface reconstruction from raw data. Incorrect normals would cause bubbles for Poisson reconstruction~\cite{10.1145/2487228.2487237}, while noisy or non-uniform points lead to incomplete meshing for ball-pivoting~\cite{817351} or Voronoi-based triangulation~\cite{boltcheva2017surface}.
}
\end{figure}

Point cloud is one of the most convenient and faithful forms captured from the real-world while mesh is a typical user-friendly 3D representation for graphics and AR/VR applications. Reconstructing mesh surfaces from points is thus a fundamental and long-standing topic~\cite{817351, 10.5555/1281957.1281965, 10.1145/2487228.2487237, boltcheva2017surface, 8954349, 10.1145/3386569.3392415}. 
However, raw point clouds produced by LiDAR sensors are often redundant, noisy, and non-uniform, which would hinder other applications such as surface reconstruction or direct rendering. 
Fig.~\ref{reconstruct} shows that direct reconstruction from such dense points can not obtain high-quality meshes.
Point cloud simplification is thus important for efficient 3D computation by reducing redundancy and optimizing quality, saving storage and transmission. Classical simplification methods~\cite{10.5555/602099.602123, 10.1145/1882261.1866190, 8101025}, which are capable of producing edge-enhanced results, cannot remedy severe noises in the input data, since only the manifold itself is exploited. 
Recently, some learning based point cloud downsampling methods are developed, which focus on classification, retrieval, and registration, etc. These algorithms do not care about the content and distribution of simplified results, which cannot be applied to surface reconstruction and rendering. The point cloud reconstruction task in~\cite{2019Learning, 9157510} aims to reconstruct dense points, which are not faithful and uniform to reconstruct high-quality surface mesh.

\begin{figure}[!tb]
\centering
\includegraphics[width=.98\linewidth]{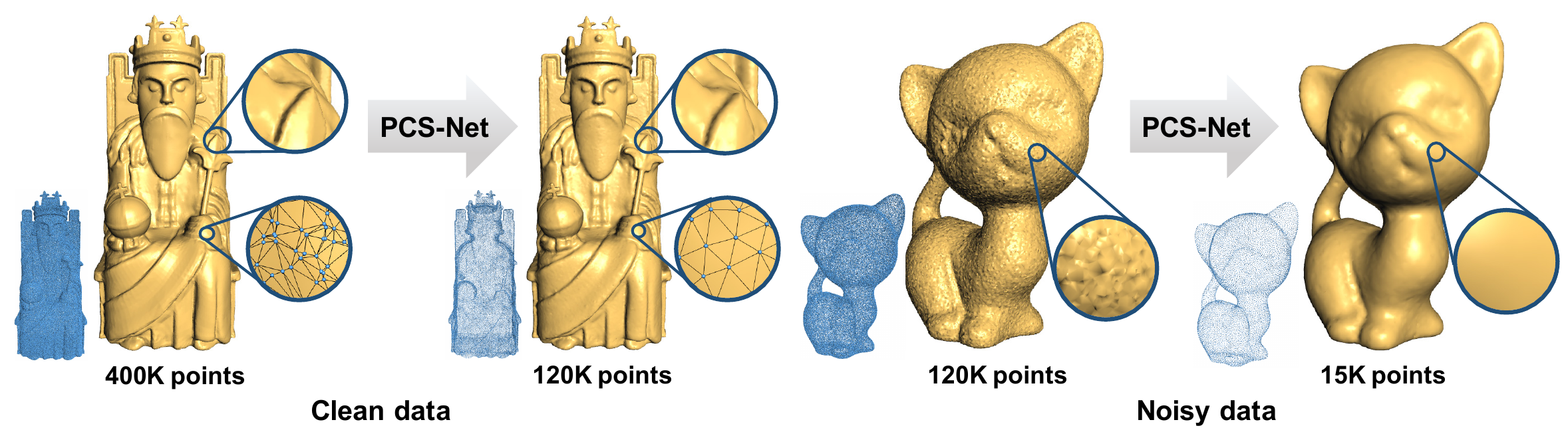}
\parbox[t]{.98\columnwidth}{\relax
}
\caption{\label{teaser}
Our PCS-Net learns to simplify point clouds, that are either clean or noisy, to reconstruct faithful surfaces accompanied with high-quality triangular meshes. 
}
\end{figure}

We propose a new point cloud simplification network (PCS-Net) to produce sparse and uniform point clouds to take advantage of compact storage and reconstruct surfaces faithfully. 
Benefiting from the fidelity and uniformity, point cloud processed by our network can be triangulated directly to produce high-quality surface meshes. 
Our PCS-Net for surface reconstruction contains two main parts. One is the \emph{feature-aware simplification module}, and the other is the \emph{double-scale resampling module}. The feature-aware simplification module learns a differentiable sampling matrix from the extracted features to simplify the input point cloud to a user-specified size. However, the resulting point cloud, which can be considered as a subset of the input, would retain defects of the input data, such as non-uniformity or noise. 
Therefore, we propose a novel double-scale resampling module to refine the positions of points. Features extracted from the dense input data are utilized as supplementary information to combine with features extracted from the simplified points. 
In this procedure, the sampling matrix learned before serves as an attention module to highlight salient features from the dense input. To further maintain shape features, we propose an adaptive sampling strategy equipped with a novel saliency loss. 
Point clouds can be simplified by our method in a feature-aware manner, i.e., points near salient features are consolidated but still with uniform distribution locally. 
Fig.~\ref{teaser} shows two simplification examples by using our method, where the input can be quite large and noisy.
In summary, our main contributions are as follows:
\begin{itemize}
\item We present a novel point cloud simplification network (PCS-Net) for surface reconstruction which, to the best of our knowledge, is the first deep learning based method for compacting points with better distribution from raw data.
\item We propose a feature-aware simplification module and a double-scale resampling module to simplify point clouds, by learning shape features and uniform distribution.
\item We design a new adaptive sampling strategy together with a novel saliency loss to maintain surface details.
\end{itemize}

\section{Related work}
\label{sec:relatedWork}

\subsection{Point cloud simplification and resampling}
Simplification and resampling are valuable for dealing with point sampled surfaces. The operators, such as locally optimal projection operator (LOP)~\cite{10.1145/1276377.1276405}, weighted LOP (WLOP)~\cite{10.1145/1661412.1618522}, and feature-preserving LOP (FLOP)~\cite{LIAO2013861}, are usually based on low geometric properties (normal vector and curvature) of point clouds. \"Oztireli \textit{et al.}~\cite{10.1145/1882261.1866190} propose a spectral analysis approach for the points simplification and resampling on manifolds. Chen \textit{et al.}~\cite{8101025} introduce a graph based algorithm for contour-enhanced resampling. Huang \textit{et al.}~\cite{10.1145/2421636.2421645} propose an edge-aware resampling algorithm. However, these traditional methods need normals of the original point cloud as input, which is an additional requirement for users. Furthermore, computing normals by fitting a local plane is vulnerable to noise. By contrast, our method only requires 3D coordinates in Euclidean space of the input data and is able to handle noisy data.

In recent years, many deep learning based methods~\cite{8099499, NIPS2017_d8bf84be, NEURIPS2018_f5f8590c, 10.1145/3326362, zhao2021point} are proposed to deal with the unordered point cloud data, inspiring a variety of novel applications and extensions~\cite{8491026, 8953650, 9156338, 8578127}. PointNet~\cite{8099499} is the pioneer network while PointNet++~\cite{NIPS2017_d8bf84be}, PointCNN~\cite{NEURIPS2018_f5f8590c}, and DGCNN~\cite{10.1145/3326362} are several popular backbones for direct point set learning. 
There are several pieces of research on learning based point set downsampling. Yang \textit{et al.}~\cite{8954194} adopt a Gumbel softmax layer to combine high level features for higher classification accuracy. Nezhadary \textit{et al.}~\cite{2020Adaptive} use critical points invoked in max-pooling as sampled points. S-Net~\cite{2019Learning} and SampleNet~\cite{9157510} generate 3D coordinates of fewer points by fully-connected layers after feature extraction. These downsampling methods could generate fewer points for higher performance than random sampling in scene understanding tasks such as classification, retrieval, and registration. The simplified point cloud obtained by these methods can also be reconstructed to dense points similar to the input, however, the simplified point cloud is a codeword of the input to some extent, which cannot retain the underlying shape faithfully. Our PCS-Net can obtain results with high fidelity and uniform point distribution, which can reconstruct precise surfaces directly using the simplified point sets. Furthermore, designing point correlation to satisfy a specific spectrum also attracts researchers’ attention~\cite{10.1145/2508363.2508375, 10.1145/2980179.2982435, 10.1145/1964921.1964945, 10.1145/2185520.2185572, 10.1145/3355089.3356562}. 

\subsection{Point cloud upsampling}
Early optimization based point cloud upsampling algorithms resort to shape priors~\cite{1175093, 10.1145/1276377.1276405, 10.1145/2601097.2601172, 10.1145/2421636.2421645}. With the development of utilizing deep networks in dealing with point clouds, some learning based point cloud upsampling methods appeared in recent years~\cite{wu2019point, 9577901, 9351772, qian2020pugeo, 9578328}. Yu \textit{et al.}~\cite{8578393} propose the PU-Net, which is the first deep network for generating a denser and uniform point cloud from a sparser set of points. PU-Net operates on patch level and expands features to obtain the result with an integer upsampling ratio. The follow-up work, EC-Net~\cite{Yu_2018_ECCV}, facilitates the consolidation of point clouds deliberately for edges. EC-Net uses a point-to-edge loss, which requires a rather expensive edge annotation for training. By contrast, our saliency loss uses training data calculated geometrically, which does not require any annotation. MPU~\cite{8953789} uses multiscale skip connections to combine local and global features, which are also used in our PCS-Net. Different frameworks such as the generative adversarial network (GAN)~\cite{9008773}, the graph convolutional network (GCN)~\cite{wu2019point, 9577901}, and the meta-learning~\cite{9351772} are applied to point set upsampling. Qian \textit{et al.}~\cite{qian2020pugeo} introduce a geometric-centric neural network, called the PUGeo-Net, to generate new samples around the input points.

\begin{figure*}[!t]
\centering
\includegraphics[width=.96\linewidth]{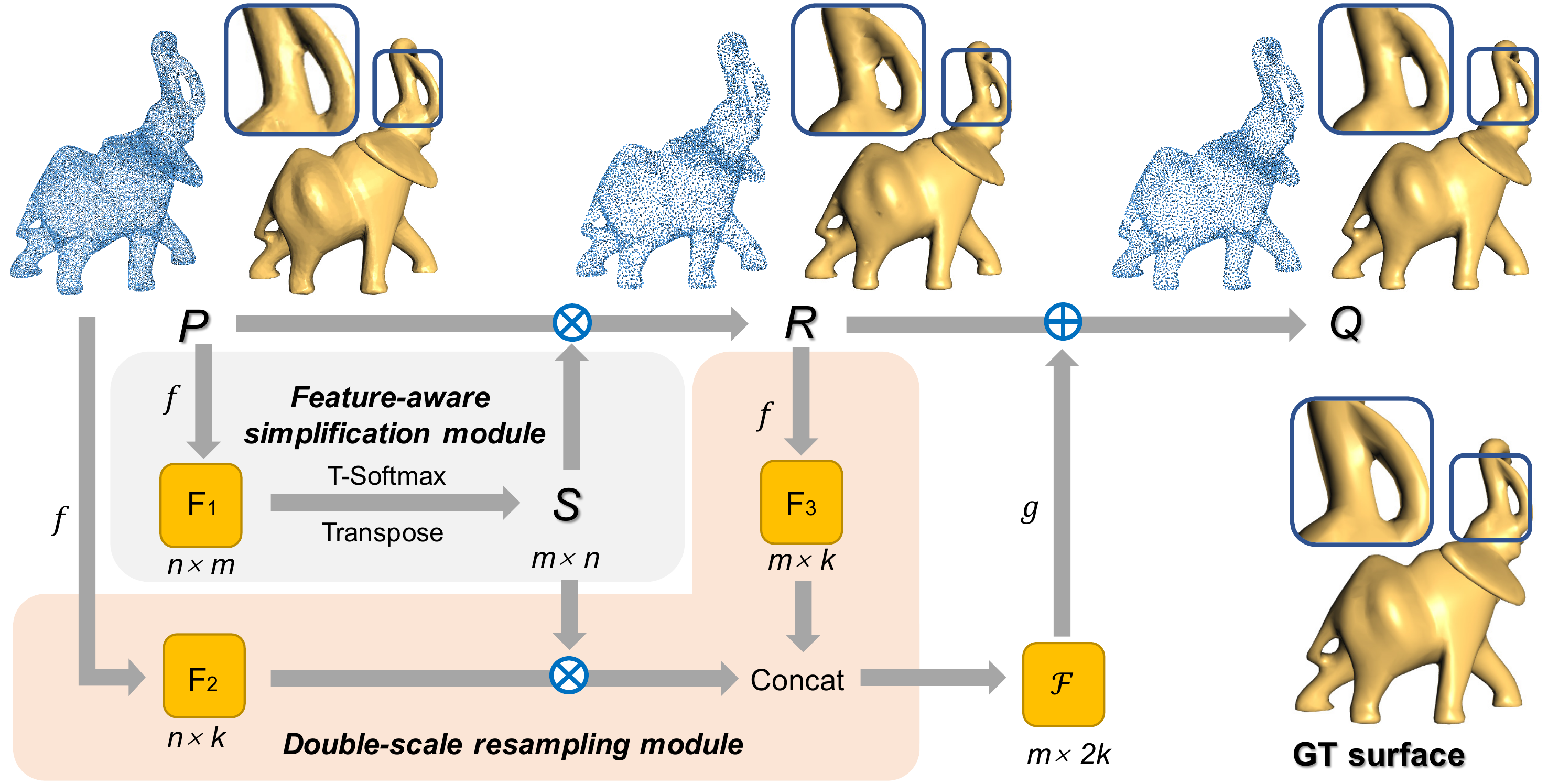}
\parbox[t]{.98\columnwidth}{\relax
}
\caption{\label{net}
The architecture of our PCS-Net contains two specially-designed modules. The \emph{feature-aware simplification module} samples fewer points by learning a sampling matrix $S$. Given the input point cloud $P$, PCS-Net first extracts a feature matrix by a feature extraction module $f$. Then we process the matrix by a softmax layer incorporating with a temperature coefficient, which is denoted as \emph{T-Softmax}, to obtain $S$. The simplified point cloud $R$ is obtained by multiplying $P$ and $S$. In the \emph{double-scale resampling module}, features are extracted from dense and sparse point clouds ($P$ and $R$), and concatenated to form a multi-scale feature matrix $\mathcal{F}\in \mathbb{R}^{m\times 2k}$, where we use $k=264$ suggested by~\cite{8953789}. Finally, the coordinate offset is regressed by a shared multilayer perceptron (MLP) layer $g$, and added with $R$ to generate the result. For comparison, we show the reconstructed surfaces of $P$, $R$, and $Q$ on the right of each point cloud.
}
\end{figure*}

\section{Method}
\subsection{Problem formulation}
Given a 3D point cloud $P = \{\mathbf{p}_i \in \mathbb{R}^3, i=1,2,...,N\}$, we aim to learn a new simplified point cloud $Q = \{\mathbf{q}_j \in \mathbb{R}^3, j=1,2,...,M\}$, of which the size is $M \leq N$. For the sake of fidelity and high-quality of the reconstructed surface, $Q$ not only fits the underlying surface faithfully but also distributes uniformly to generate meshes that are as close to regular triangles as possible. 
Thus the goal of this paper is to obtain a point cloud with fewer points, from which we can reconstruct a faithful and higher-quality surface mesh that tries to satisfy the following formulation: 
\begin{eqnarray}
&\min \Vert \mathcal{S}_{un}(P)-\mathcal{S}(Q) \Vert  \quad \quad \text{s.t. $Q=\mathcal{T}(P)$}, \\
&\min \sum | \mathcal{A}(\mathcal{S}(Q))-\frac{\pi}{3} |  \quad \quad \text{s.t. $Q=\mathcal{T}(P)$},
\end{eqnarray}
where $\mathcal{T}(\cdot)$ denotes our network, $\mathcal{S}_{un}(\cdot)$ is the underlying surface which is represented by the user acquired point cloud $P$, $\mathcal{S}(\cdot)$ represents the triangular surface mesh reconstructed from the point cloud, and $\mathcal{A}(\cdot)$ represents the inner angle of the triangles in the mesh. 

\subsection{Network architecture}
\label{3.2}
We present a novel PCS-Net based on the core idea of learning a \textit{sampling matrix} to determine which point in the original point cloud should be sampled and then \textit{refining} their positions to remove defects in raw data. However, since the number of points in $P$ is usually large, directly processing $P$ would lead to GPU out of memory. 
To overcome this issue, inspired by the reconstruction-oriented point cloud upsampling methods~\cite{8578393, 8953789, 9008773, qian2020pugeo}, we preprocess the input dense point cloud by partitioning it into smaller patches.
The patch seeds are first sampled from $P$ by using the farthest point sampling (FPS) algorithm. Then we generate patches by computing the Voronoi diagrams corresponding to the seeds. 

Without loss of generality, our end-to-end neural network which processes one patch $P^\prime$, is illustrated in Fig.~\ref{net}. 
Given the input point cloud, we start by the \emph{feature-aware simplification module} that returns a simplified point cloud with the user-specified size. This module is essentially an attention learning process to select points from input points:
\begin{eqnarray}
\label{r}
R^\prime=SP^\prime,
\end{eqnarray}
where $R^\prime \in \mathbb{R}^{m \times 3}$ is the simplified point cloud, $S\in \mathbb{R}^{m\times n}$ represents the learned sampling matrix, $n$ and $m$ denote number of points in $P^\prime$ and $R^\prime$. We intend to obtain a subset point cloud $R^\prime \subseteq P^\prime$, thus each row of the sampling matrix $S$ is required to be a one-hot vector, elements of which contain only one $1$ and the rest are $0$. However, the one-hot matrix cannot be directly used in a network due to the non-differentiable property. Therefore, $S$ is an approximate one-hot matrix which will be detailed below.

To learn such a sampling matrix, a feature extraction component, denoted as $f$ in Fig \ref{net}, first extracts the feature matrix $F_1$ from the input point cloud hierarchically, following the upsampling method MPU~\cite{8953789}, which is a lightweight and efficient backbone for dealing with point sets. We also adopt the dense connection and feature-based k-nearest neighbors (kNN) to extract and skip-connect features hierarchically.

To obtain an approximate one-hot sampling matrix, the temperature-softmax layer (T-Softmax in Fig.~\ref{net}) is applied to the feature matrix $F_1$. 
The T-Softmax can force each row of the matrix to approach a one-hot vector, such that the simplified point set would not deviate from the underlying shape too far. We define this process as:
\begin{eqnarray}
F_1=\Big[[a_{11},...,a_{n1}]^{\top},...,[a_{1m},...,a_{nm}]^{\top}\Big],\\
\begin{aligned}
\label{t}
S=\Big[[\frac{e^{a_{11}}/t^2}{\sum_{i=1}^n e^{a_{i1}}/t^2},...,
\frac{e^{a_{n1}}/t^2}{\sum_{i=1}^n e^{a_{i1}}/t^2}]^{\top},...,\\
[\frac{e^{a_{1m}}/t^2}{\sum_{i=1}^n e^{a_{im}}/t^2},...,
\frac{e^{a_{nm}}/t^2}{\sum_{i=1}^n e^{a_{im}}/t^2}]^{\top}\Big]^{\top},
\end{aligned}
\end{eqnarray}
where $a_{11},...,a_{nm}$ denote elements in the feature matrix $F_1$, $t$ is a temperature coefficient. When $t \rightarrow 0$, each row of $S$ converges to a one-hot vector, and $R^\prime$ becomes a subset of $P^\prime$. We set $t$ to $1$ at the beginning of the training, and this coefficient is gradually reduced to $0.1$. The implementation details will be explained in Section \ref{detail}.

Since point sets acquired by 3D scanners are often non-uniform and noisy, directly multiplying the input point set $P^\prime$ by the sampling matrix $S$ would retain these defects. These defects lead to some annoying artifacts, such as deviation on some complex structures and obvious masses on smooth surfaces which are shown in the surface reconstructed from $R$ in Fig.~\ref{net}. Therefore, we propose a novel \emph{double-scale resampling module} to refine the positions of the sampled points, to achieve high fidelity and uniform distribution. We extract features ($F_2$ and $F_3$) from both the dense point cloud $P^\prime$ and the sparse one $R^\prime$ by using the same $f(\cdot)$ as the simplification module. The feature matrix $F_2$ is first multiplied by the sampling matrix $S$. Then we concatenate it with the feature extracted from the sparse one, in order to preserve more details of the underlying surface. The combined features from these double scales are fed to a shared MLP layer of size $[128, 128, 64, 3]$ to regress a coordinate offset, which is added to $R^\prime$ to generate the point cloud $Q^\prime$. 

In summary, the overall network can be formulated as:
\begin{eqnarray}
\label{networkf}
R^\prime&=&T\text{-}Softmax\Big(f(P^\prime)\Big)P^\prime , \\
\label{networkdouble}
\mathcal{F}&=&Concat\Big(f(P^\prime),f(R^\prime)\Big) ,\\
\label{networkg}
Q^\prime &=&R^\prime+g(\mathcal{F}),
\end{eqnarray}
where $T\text{-}Softmax(\cdot)$ denotes the softmax layer with temperature coefficient in Eq.~\ref{t}, $f(\cdot)$ represents the feature extraction module and $g(\cdot)$ is the shared MLP generating coordinate offsets. Finally, our simplified point cloud $Q$ is constituted of $Q^\prime$ of each patch.

\noindent\subsection{Loss functions}
Our network training is supervised by an efficient joint loss function, which is specially designed for our PCS-Net to obtain faithful and uniform points:
\begin{eqnarray}
\label{L}
L=\alpha L_{r}+ L_{sp}+\beta L_{rep}.
\end{eqnarray}
Our loss $L$ is the sum of a reconstruction term $L_{r}$ to maintain points near the underlying surface, a spread term $L_{sp}$ to ensure the points consolidate near features and spread over the surface, and a repulsion term $L_{rep}$ to generate uniform distribution in local. We set $\alpha=0.01$ and $\beta=0.0012$ according to the experimental experience.

\noindent\textbf{Reconstruction loss} restricts the distance of output points from the input ones, which facilitates the results near the underlying surface. It is defined as following to maintain the shape fidelity:
\begin{eqnarray}
\label{Lr}
L_{r}=\frac{1}{m} \sum_{i=1}^m \min_{\mathbf{x}\in P^\prime}\Big\Vert \mathbf{y_i} - \mathbf{x} \Big\Vert ^2 ,
\end{eqnarray}
where $\mathbf{x}$ and $\mathbf{y}_i$ denote the 3D coordinates of one point in $P^\prime$ or $Q^\prime$.

\noindent\textbf{Spread loss} forces the points to spread over the input dense point cloud. It can be formulated as:
\begin{eqnarray}
\label{Lsp}
L_{sp}=\frac{1}{n} \sum_{j=1}^n \min_{\mathbf{y}\in Q^\prime} \Big\Vert \mathbf{x_j} - \mathbf{y} \Big\Vert ^2.
\end{eqnarray}

\noindent\textbf{Repulsion loss} is used to make the generated points distribute more uniformly in local regions. Globally, points would gather near the salient features of the surface. Locally, uniformly distributed points would facilitate the reconstruction of high-quality meshes. The repulsion loss is formulated to maximize the distance between any two neighboring points in the output cloud $Q^\prime$:
\begin{eqnarray}
L_{rep}=-\frac{1}{m}\sum_{j=1}^m { \min_{\mathbf{y}\in Q^\prime , \mathbf{y}\neq \mathbf{y_j}} \Big\Vert \mathbf{y}-\mathbf{y_j} \Big\Vert ^2}.
\end{eqnarray}

\begin{figure}[!tb]
\centering
\includegraphics[width=.98\linewidth]{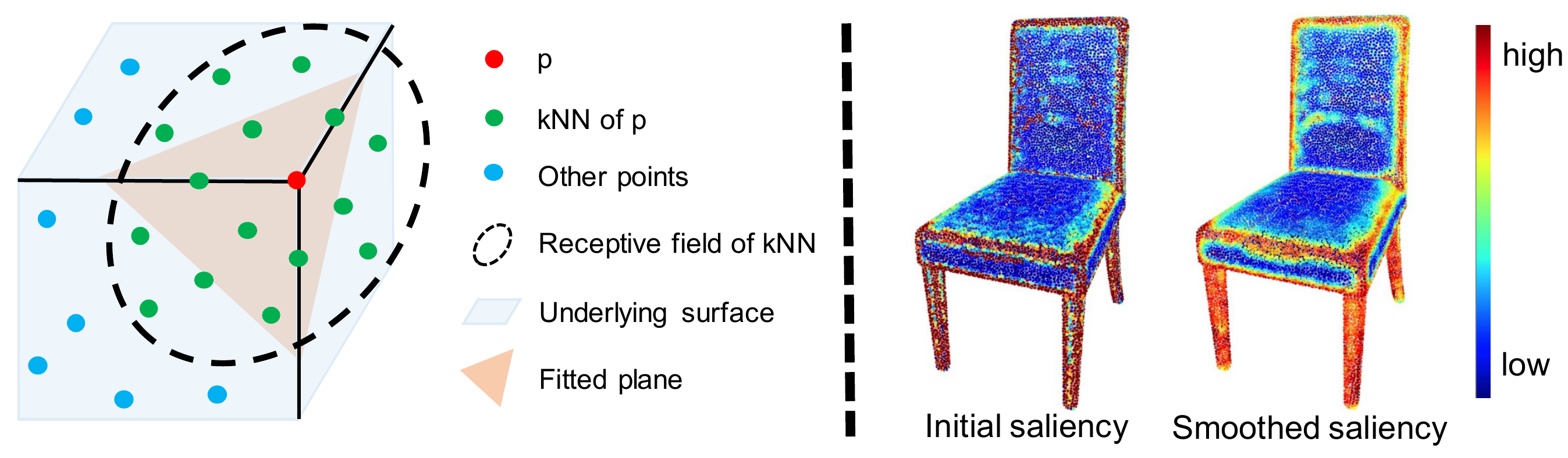}
\parbox[t]{.98\columnwidth}{\relax
}
\caption{\label{saliency}
Left: The saliency of a point $\mathbf{p}$ is computed as the Euclidean distance between the red point and the pink fitted plane. Right: A mesh example showing the initially computed saliency on it and the final smoothed saliency.
}
\end{figure}

\subsection{Adaptive sampling}
\label{Adaptive_sampling}
To preserve more details, more points are preferred to be sampled from the regions which have rich details and salient features. To achieve this goal, we propose both global and local adaptive sampling strategies to handle the shape features.

\noindent\textbf{Saliency loss.}
A quantitative indicator is needed to measure the importance of regions on the surface. We propose to use a geometry-based metric, the saliency, in our adaptive sampling strategy. Specifically, 
for one point $\mathbf{p}_i$ in the point set $P$, we use its k-nearest ($k=20$) neighbors $\xi(\mathbf{p}_i)=\{\mathbf{p}_i^1,\mathbf{p}_i^2,...\mathbf{p}_i^k\}$ to measure its saliency:
\begin{eqnarray}
\label{saliencyeq1}
s_i=D(p_i,Plane(\xi(\mathbf{p}_i))).
\end{eqnarray}
Here we first regress a plane using $\xi(\mathbf{p}_i)$ by a least-squares fitting, which is denoted as $Plane(\cdot)$. Then we calculate the saliency as the distance between $\mathbf{p}_i$ and the fitted plane, denoted as $D(\cdot)$ (see the left side of Fig.~\ref{saliency} for an illustration). 

Equipped with saliency, we then propose a saliency loss, instead of the spread loss, to force the points to not only spread over the input dense points but also consolidate near the features of the underlying surface. Our saliency loss is defined as:
\begin{eqnarray}
\label{Ls}
L_{s}=\frac{1}{n}\sum_{j=1}^n \hat{s_j}\min_{\mathbf{y}\in Q^\prime} \Big\Vert \mathbf{x_j} - \mathbf{y} \Big\Vert ^2,
\end{eqnarray}
where $\hat{s_j}$ is the saliency of the $j^{th}$ point in $P$. However, the saliency directly computed by Eq.~\ref{saliencyeq1} would be too low on the flat regions of the shape, which easily leads to hollows in the reconstructed surface. To avoid it, we smoothen $s_i$ and use the smoothed saliency $\hat{s_i}$ in the loss function (Eq.~\ref{Ls}),
\begin{eqnarray}
\label{saliencyeq2}
\hat{s_i}=\Big(\sum_{\hat {\mathbf{x}_j}\in \xi (\mathbf{x}_i)} s_j e^{\frac{-\Vert \mathbf{x}_i - \hat {\mathbf{x}_j} \Vert^2}{h^2}}\Big) / \Big(\sum_{\hat {\mathbf{x}_j}\in \xi (\mathbf{x}_i)} e^{\frac{-\Vert \mathbf{x}_i - \hat {\mathbf{x}_j} \Vert^2}{h^2}} \Big),
\end{eqnarray}
where $\mathbf{x}$ represents the coordinate, $\xi(\mathbf{x}_i)$ denotes k-nearest neighbors of $\mathbf{x}_i$. We set the smoothing parameter $h=0.01$ and $k=20$. An example model with point saliency is shown in Fig.~\ref{saliency}. Note that point saliency is only required in the training set for loss calculating. In the testing phase, only the 3D coordinates of the input points are needed.

\begin{figure}[tb]
\centering
\includegraphics[width=.6\linewidth]{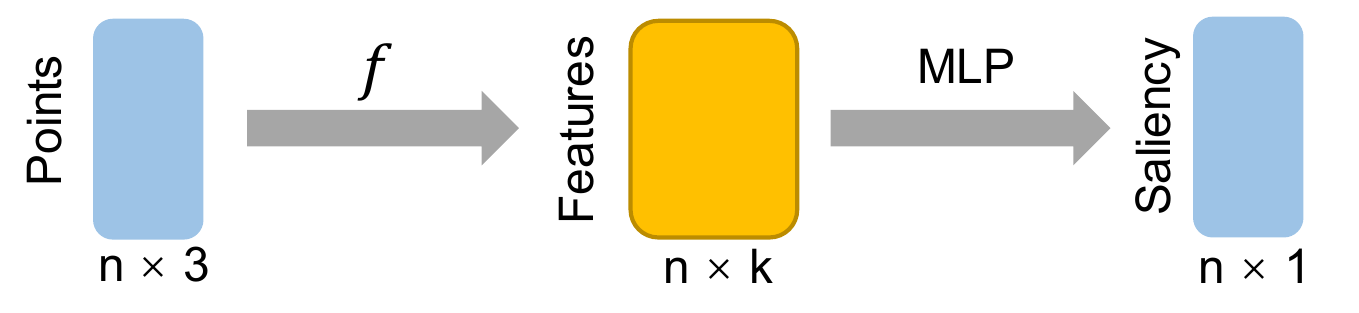}
\parbox[t]{.6\columnwidth}{\relax
}
\caption{\label{saliencynetwork}
The architecture of the saliency prediction network. From an input point cloud, a feature matrix is extracted by using the feature extraction component $f$ in Eq.~\ref{networkf}. The saliency is then regressed by a shared MLP layer of size [128,128,64,1]. 
}
\end{figure}

\begin{figure}[tb]
\centering
\includegraphics[width=1.0\linewidth]{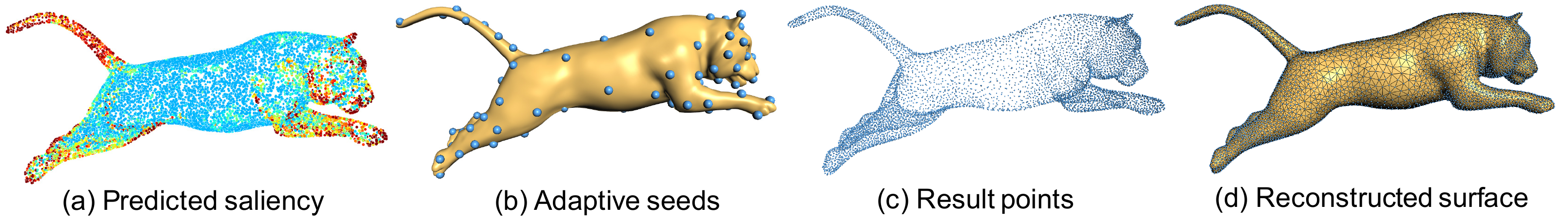}
\parbox[t]{1.0\columnwidth}{\relax
}
\caption{\label{adaptivepatch}
After predicting saliency of input points (a) where red color represents large values, we can obtain adaptive seeds (b). Then points are generated based on adaptive patches (c). Finally, vertices of the reconstructed surface consolidate near the features (d).  
}
\end{figure}

\noindent\textbf{Adaptive patches.}
We use adaptive patches by weighted Voronoi diagram to achieve global adaptive sampling. Seeds are obtained by using a 3D Poisson-disk sampling method~\cite{guo2015efficient}, of which the point weight is replaced by the saliency, which can be effectively predicted by using the network illustrated in Fig.~\ref{saliencynetwork}. 
After obtaining adaptive seeds, patches are cut by using the Voronoi diagram.

Based on those adaptive patches, our PCS-Net can generate points consolidating near the features. Finally, the mesh surface is reconstructed by using~\cite{boltcheva2017surface} which directly triangulates points. Compared to smooth-prior methods such as Poisson reconstruction~\cite{10.1145/2487228.2487237}, the algorithm of~\cite{boltcheva2017surface} would not change coordinates of points, thus the shape details, as well as defects, of the point cloud can be retained. Therefore, the quality of points plays a crucial role in this reconstruction procedure. Stages of our adaptive sampling pipeline are illustrated in Fig.~\ref{adaptivepatch}.

\begin{figure*}[!t]
\centering
\includegraphics[width=1.0\linewidth]{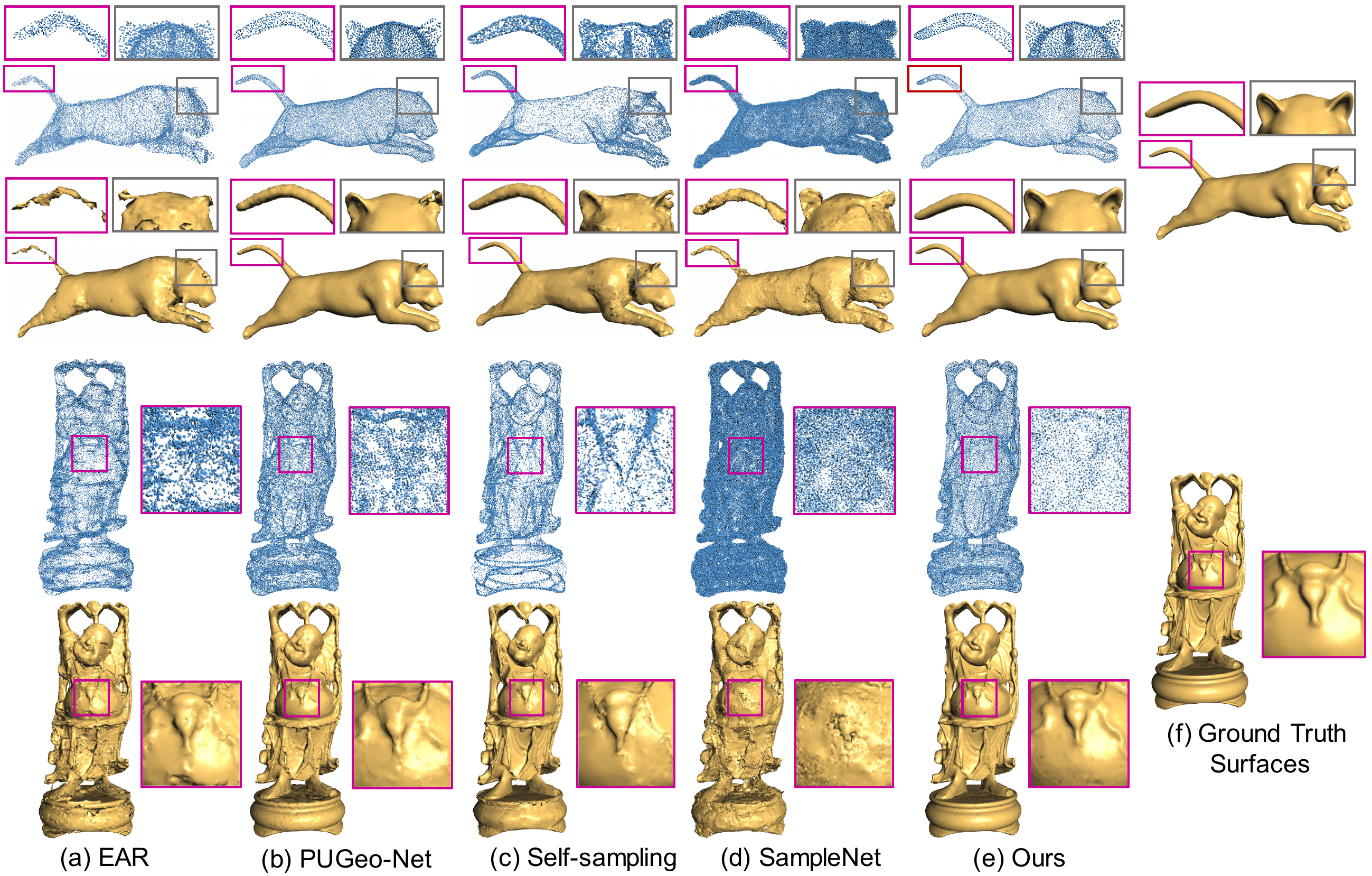}
\parbox[t]{1.0\columnwidth}{\relax
}
\caption{\label{resultmesh}
Qualitative comparison to previous methods using examples of the 'Tiger' and 'Happy buddha'. For each method, we show its output point cloud and the corresponding reconstructed surface mesh. Zoomed-in views are also presented to show detailed comparisons.
}
\end{figure*}

\section{Experimental Results}
\label{sec:results}
\subsection{Implementation details}
\label{detail}
Following previous reconstruction-oriented methods~\cite{8953789, 9008773}, we use 175 mesh models in the Visionair repository \cite{data} to construct our dataset, including both simple and complex models. 
We separate the dataset into a training set with 140 models, a testing set with 18 models, and a validation set with 17 models.
The original point cloud ($150K$ points) is randomly sampled from each model.
\begin{wrapfigure}{r}{0.18\textwidth} 
    \centering
    \includegraphics[width=0.18\textwidth]{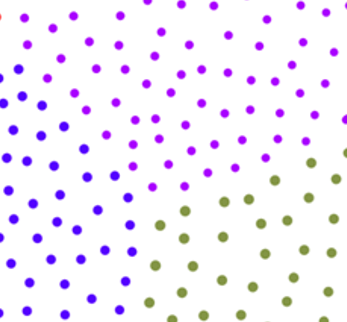}
\end{wrapfigure}
The uniform or adaptive patches are obtained by using our approaches proposed in Section \ref{3.2} or \ref{Adaptive_sampling}, where we generate 120 patches for one model.
Then in each patch, we randomly selected 1024 points to constitute input point cloud $P^\prime$ for our PCS-Net, thus the input points are non-uniform. 
At testing stage, our method outputs $15K$ points for each model. Note that we would not generate artifacts at patch boundaries because we obtain the training data, including points and saliency, before cutting patches. The inset figure visualizes the local neighboring patches of our results.


Our network is implemented in PyTorch on an NVIDIA 2080Ti graphics card. 
We train the PCS-Net for $2000$ epochs.
The temperature coefficient $t$ in Eq.~\ref{t} is $1$ before the $1200^{th}$ epoch and evenly reduce to $0.1$ from epoch $1200$ to $1600$.
The loss is calculated between output point cloud patch $Q^\prime$ and the input $P^\prime$. 
Uniform mode and adaptive mode use the same parameter settings. 
We will open-source the code and data to facilitate future research.

\subsection{Comparisons}
\label{comparisons}
To demonstrate the effectiveness our PCS-Net, we now compare it against various competitors, including a traditional resampling method (EAR~\cite{10.1145/2421636.2421645}) and the state-of-the-art downsampling networks (\textit{i.e.}, S-Net~\cite{2019Learning}, SampleNet~\cite{9157510}, Self-sampling~\cite{10.1145/3470645}, CP-Net~\cite{2020Adaptive}). Most of the downsampling methods do not focus on the surface reconstruction task~\cite{2019Learning, 9157510, 2020Adaptive}. To further reveal the influence of the point quality to surface reconstruction, we also compare with several reconstruction-oriented deep upsampling methods, (\textit{i.e.}, MPU~\cite{8953789}, PU-GAN~\cite{9008773}, PUGeo-Net~\cite{qian2020pugeo}), which use less input points to output the same size point clouds as downsampling methods.
We re-train all of the deep neural models on our training dataset.  
Due to the page limit, we only show some selected results, and more exhaustive comparisons are provided in supplemental materials.


\begin{figure}[!tb]
\centering
\includegraphics[width=.95\linewidth]{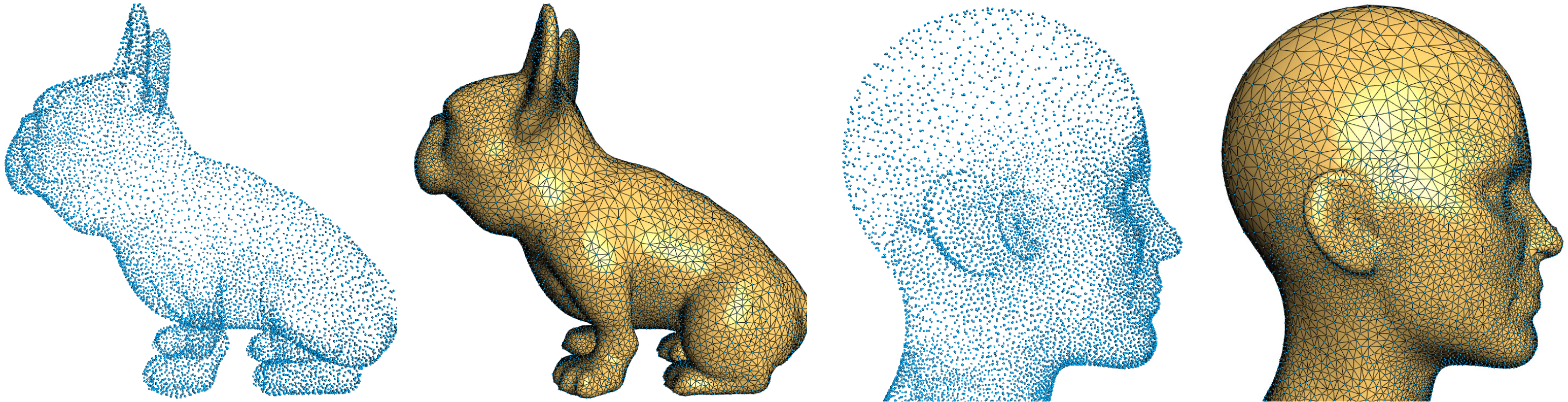}
\parbox[t]{.95\columnwidth}{\relax
}
\caption{
Adaptive sampling and surface reconstruction results using our method. Vertices and mesh lines are visualized to show the different triangle sizes in different parts. 
}
\label{showadaptive}
\end{figure}

\noindent\textbf{Qualitative comparison.}
Fig.~\ref{resultmesh} shows the qualitative comparison results, 
where the 'Tiger' model containing $150K$ points is from our test set, and the 'Happy buddha' containing $550K$ points is a real-scanned model.
The output point cloud of SampleNet~\cite{9157510} has the same point number as input due to its decoder, while the output of all other methods each contains $15K$ points for 'Tiger' and $120K$ points for 'Happy buddha'. The upsampling methods use $4\times$ upsampling ratio.
It can be seen that other methods generate more noisy and non-uniform point distributions, leading to crack or zigzag artifacts on the reconstructed surfaces. 
By contrast, our PCS-Net produces results with higher fidelity that is shown in the zoomed-in views. 
To be specific, SampleNet~\cite{9157510} is skilled in maintaining semantic information but not geometry. Self-sampling~\cite{10.1145/3470645} can consolidate points near edges but produce noise (\textit{e.g.} the decoration of 'Happy buddha') and large non-uniform regions (\textit{e.g.} head of 'Tiger'). PUGeoNet~\cite{qian2020pugeo} can generate more smooth and faithful results than EAR and SampleNet. However, it also suffers from non-uniformity (\textit{e.g.}, the decoration of 'Happy buddha') or outliers and deficiencies (\textit{e.g.}, right ear of 'Tiger'). Our PCS-Net is able to obtain results without these artifacts to achieve the most faithful surfaces.
To be fair, we only compare to previous approaches using the uniform mode of our method. Besides, Fig.~\ref{showadaptive} demonstrates the ability of our PCS-Net to produce adaptive point distributions for preserving salient features.

\begin{figure}[!tb]
\centering
\includegraphics[width=.95\linewidth]{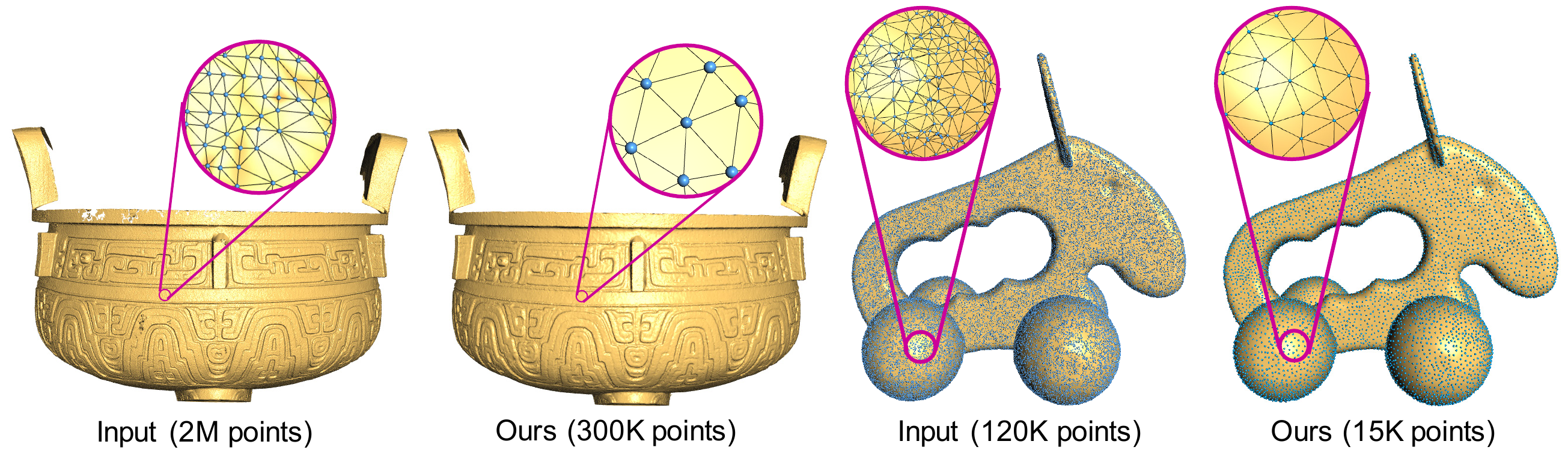}
\parbox[t]{.95\columnwidth}{\relax
}
\caption{
Our PCS-Net can generate uniform points from a real-scanned input (left) or a non-uniform input (right). 
}
\label{uniformPoints}
\end{figure}

Furthermore, compared with the input non-uniform points, our PCS-Net generates results with uniform distribution to reconstruct high-quality triangular meshes. Fig.~\ref{uniformPoints} shows a real-scanned input point cloud (top) and a random distribution input (bottom), which are simplified by our method to uniform results to reconstruct compact high-quality meshes. 

\begin{table}[!tb]
\centering
\caption{Quantitative comparison of different methods in terms of geometric fidelity and mesh quality on our testing dataset. The best results are marked in \textcolor{myred}{red} color, and the second best results are in \textcolor{myblue}{blue} color.}
\label{Quantitative}
\begin{tabular}{c|c|c|c|c|c|c}
\hline
              Methods & $D_c (e^{-3}) \downarrow$  & $D_h (e^{-2}) \downarrow$ & $p2f (e^{-4})$ & $G \uparrow$  & $\theta_{avg} \uparrow$ & $\%_{<30^{\circ}} \downarrow$\\
\hline
  EAR~\cite{10.1145/2421636.2421645} & 12.9 &5.40 &30.3 &0.788 &43.2 &7.51  \\  
  MPU~\cite{8953789}              & 4.24 &1.08 &9.40 &0.771 &42.3 &6.09  \\
  PU-GAN~\cite{9008773}        & 5.61 &3.75 &19.7 &0.775 &42.5 &5.52 \\
  PUGeo-Net~\cite{qian2020pugeo} & \textcolor{myblue}{4.05} &1.16 &\textcolor{myblue}{6.47} &\textcolor{myred}{0.794} &\textcolor{myred}{44.0} &\textcolor{myblue}{5.25} \\
  S-Net~\cite{2019Learning}  & 11.4 &5.48 &25.9 &0.715 &36.9 &26.4      \\
  CP-Net~\cite{2020Adaptive}        & 35.1 &12.6 &37.6 &0.719 &39.1 &19.5   \\
  SampleNet~\cite{9157510}     & 9.80 &4.51 &19.1 &0.708 &36.3 &28.1      \\
  Self-sampling~\cite{10.1145/3470645} & 5.08 &\textcolor{myblue}{0.81} &8.62 &0.714 &38.3 &20.6   \\
  Ours          & \textcolor{myred}{3.90} &\textcolor{myred}{0.74} &\textcolor{myred}{3.73} &\textcolor{myblue}{0.790} &\textcolor{myblue}{43.6} &\textcolor{myred}{4.32} \\
\hline
\end{tabular}
\end{table}

\noindent\textbf{Quantitative comparison.}
To quantitatively compare our PCS-Net with other methods, we introduce mesh distance as a quantitative metric for measuring the approximation error between the results and the ground truth model. We randomly sample dense point clouds $P_D$ and $Q_D$ (both containing $W=100K$ points) from the ground truth model and the reconstructed surface using the method of~\cite{6143943}. The mesh distance between $P_D$ and $Q_D$ includes two terms, the Chamfer distance and the Hausdorff distance:
\begin{eqnarray}
\begin{aligned}
\label{dc}
D_{c}=\frac{1}{W} \sum_{i=1}^W {\min_{\mathbf{v}\in Q_D} {\Vert \mathbf{u_i}-\mathbf{v} \Vert ^2}} 
+ \frac{1}{W} \sum_{j=1}^W {\min_{\mathbf{u}\in P_D} \Vert \mathbf{v_j}-\mathbf{u} \Vert ^2},
\end{aligned}
\end{eqnarray}
\begin{eqnarray}
\begin{aligned}
\label{dh}
D_{h}=\max {\Big( \max_{\mathbf{u}\in P_D} \min_{\mathbf{v}\in Q_D} {\Vert \mathbf{u}-\mathbf{v} \Vert ^2}, 
\max_{\mathbf{v}\in Q_D} \min_{\mathbf{u}\in P_D} {\Vert \mathbf{v}-\mathbf{u} \Vert ^2}  \Big)},
\end{aligned}
\end{eqnarray}
where $\mathbf{u}_i$ or $\mathbf{v}_j$ is one point in $P_D$ or $Q_D$, respectively. 
We also use mean of point-to-face distance ($p2f$) between the output points and ground truth meshes to directly evaluate the fidelity.
Moreover, since we aim to obtain high-quality surface reconstruction, we use the metrics introduced in~\cite{Frey99} to evaluate the quality of mesh elements.
In particular, the quality of a triangle mesh can be measured by $G=2\sqrt{3}\frac{S}{ph}$,
where $S$ is the triangle area, $p$ represents the half-perimeter, $h$ denotes the length of the longest edge. $G\in[0,1]$, and $G=1$ denotes a regular triangle. 
Besides, we also consider the triangle angles, where $\theta_{avg}$ denotes the average of the minimum angles in all triangles, and $\%_{<30^{\circ}}$ represents the percentage of triangles with their minimal angles smaller than $30$ degrees.
The numerical statistics of different methods on our test dataset are reported in Table~\ref{Quantitative}.
From the comparison, we can see that our PCS-Net achieves the best performance for maintaining the geometric fidelity of the shape. 
The PUGeo-Net~\cite{qian2020pugeo}, a state-of-the-art upsampling method, has the best uniform point distribution (using the metrics of $G$ and $\theta_{avg}$), but ours is still competitive with it while our $\%_{<30^{\circ}}$ is much better than it.

\begin{figure}[tb]
\centering
\includegraphics[width=.98\linewidth]{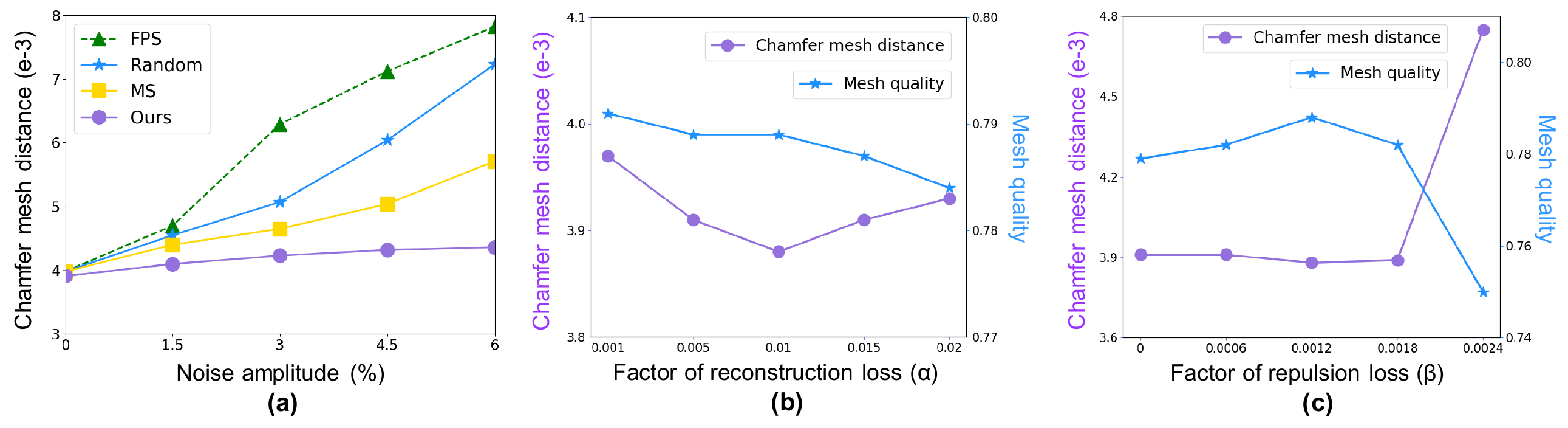}
\parbox[t]{.98\columnwidth}{\relax
}
\caption{\label{noise}
(a) Comparison with FPS, random sampling and mesh simplification (MS) in terms of fidelity on different noise amplitudes. The error of FPS is even larger than random sampling because FPS tends to select outliers. (b, c) Impact of parameter selection on fidelity and uniformity. The results are obtained in uniform mode. 
}
\end{figure}

In addition, compared to traditional methods of directly selecting points from the input, \textit{e.g.}, farthest point sampling (FPS) or random sampling, our learning based method would not be seriously affected by noise. In Fig.~\ref{noise}(a), we compute the Chamfer mesh distance under different levels of noises to demonstrate that our PCS-Net is more robust to noise than FPS and random sampling. 

Finally, an intuitive way to obtain a compact surface is first reconstructing the 3D mesh from dense raw data then simplifying it. We compare PCS-Net to such a mesh simplification pipeline, which directly applies screened Poisson reconstruction~\cite{10.1145/2487228.2487237} and simplifies the results using~\cite{809869}. 
Fig.~\ref{noise}(a) shows the comparison of different methods on various noise amplitudes. The noise amplitude denotes the standard deviation, which is measured by the percentage of the bounding sphere's radius, of Gaussian noise. Note that we use clean data to calculate the training loss functions. And we only train the network once using $6\%$ noisy input, results of different noise amplitudes are all predicted by the same trained model. This comparison reveals that although the Poisson reconstruction method can decrease the noise of input data, our PCS-Net still achieves better results than it. 

\subsection{Ablation study}
We first conduct an ablation study towards our network to evaluate the contribution and effectiveness of our proposed double-scale resampling module and the adaptive strategy. 
Table~\ref{Ablation} lists the evaluation results, where the baseline approach only adopts the simplification module, and the uniform mode uses the double-scale resampling module but without the adaptive strategy. As it is shown, the baseline would retain defects of the input data, leading to distortion and rough surface, which causes high approximation error and low uniformity (see Fig.~\ref{net} for a visual illustration). By comparison, using the double-scale resampling module or the adaptive strategy would both improve the point cloud quality. However, there is still a trade-off between fidelity and uniformity in our adaptive strategy.
The adaptive strategy could reduce the approximation error due to points consolidating at the regions with rich shape details to retain these features, but loses a fraction of uniformity. The adaptive sampling results, which maintain salient features by generating more points in these regions, are shown in Fig. \ref{showadaptive}.


\begin{table}[!t]
\begin{center}
\caption{
Ablation study of our proposed double-scale resampling module and adaptive strategy. The best result of each measurement is marked in \textbf{bold} font.}
\label{Ablation}
\begin{tabular}{c|c|c|c|c|c|c}
\hline
    Modes          & $D_c (e^{-3}) \downarrow $  & $D_h  (e^{-2}) \downarrow$ & $p2f (e^{-4}) \downarrow$  & $G \uparrow$  & $\theta_{avg} \uparrow$ & $\%_{<30^{\circ}} \downarrow$\\
\hline
  Baseline           & 4.22 &1.03 & 5.72 &0.752 &40.5 &14.3      \\  
  Uniform     & 3.90 &0.74 &3.73 &\textbf{0.790} &\textbf{43.6} &\textbf{4.32}      \\
  Adaptive     & \textbf{3.86} &\textbf{0.73} &\textbf{3.71} &0.789 &43.4 &5.01     \\
\hline
\end{tabular}
\end{center}
\end{table}


We also evaluate the terms of our loss function by analyzing their parameters. Fig.~\ref{noise} (b) and (c) show the impact of selecting different parameters. We can see that appropriate reconstruction loss can achieve high fidelity, while appropriate repulsion loss forces points to distribute uniformly to reconstruct high-quality meshes. In all of our experiments, we use $\alpha=0.01$ and $\beta=0.0012$ to compromise the reconstruction fidelity and mesh quality.



\section{Conclusion}
\label{sec:conclusion}

We have presented a novel point cloud simplification method for high-quality surface reconstruction. The core of our method is the PCS-Net, which contains a feature-aware simplification module and a double-scale resampling module. Points are simplified by a learned matrix and refined by using features extracted from dense and sparse scales to obtain faithful and uniform results. 
An adaptive resampling strategy, together with a novel saliency loss is developed to reconstruct details faithfully. 
Both qualitative and quantitative experiments verify our superior performance over state-of-the-art methods.

\clearpage
%
%
\bibliographystyle{splncs04}
\bibliography{egbib}
\end{document}